\documentclass[letterpaper]{article} 
\usepackage{aaai25}  
\usepackage{times}  
\usepackage{helvet}  
\usepackage{courier}  
\usepackage[hyphens]{url}  
\usepackage{graphicx} 
\usepackage{natbib}  
\usepackage{caption} 
\usepackage{algorithm}
\usepackage{algorithmic}
\usepackage{amsmath}
\nocopyright
\usepackage{newfloat}
\usepackage{listings}
\usepackage{multirow}
\usepackage{booktabs}
\usepackage{xcolor}
\usepackage{colortbl}
\usepackage{subcaption}
\DeclareCaptionStyle{ruled}{labelfont=normalfont,labelsep=colon,strut=off} 
\floatstyle{ruled}
\newfloat{listing}{tb}{lst}{}
\floatname{listing}{Listing}
\title{PromptTea: Let Prompts Tell TeaCache the Optimal Threshold}

\author{
    Zishen Huang\textsuperscript{\rm 1},
    Chunyu Yang\textsuperscript{\rm 1} \thanks{Correspondence author}, 
    Mengyuan Ren\textsuperscript{\rm 1},
}
\affiliations{
    \textsuperscript{\rm 1}Ucap Cloud Information Technology Co.,Ltd \\
    huangzs@ucap.com.cn
}

\usepackage{bibentry}

\begin{document}

\maketitle
\begin{figure*}[t]
\centering
    \begin{subfigure}[b]{0.42\textwidth}
        \includegraphics[width=\textwidth]{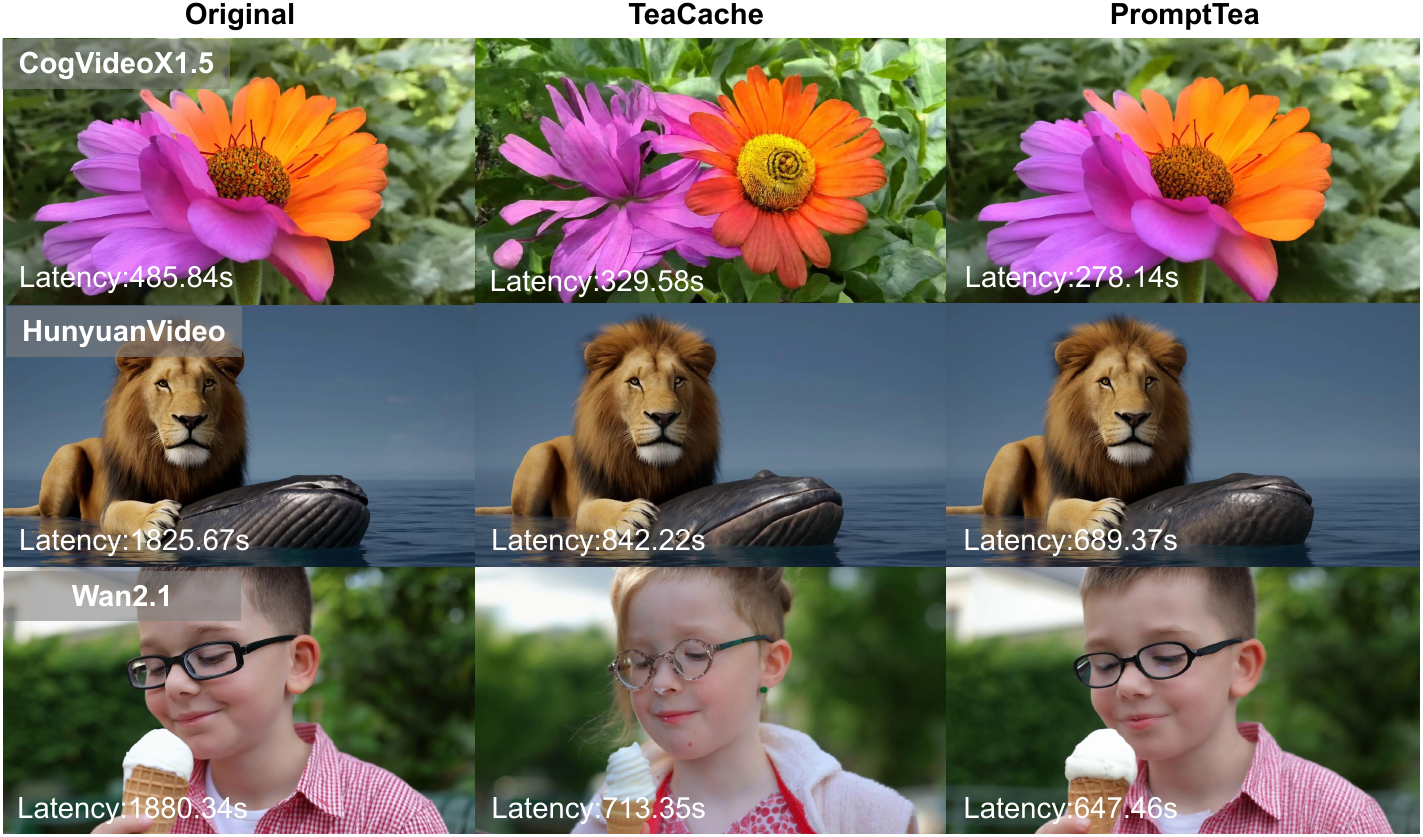}
        \caption{Visual Quality}
        \label{fig.Visual Quality}
    \end{subfigure}
    \hspace{0.05\textwidth}
    \begin{subfigure}[b]{0.3\textwidth}
        \includegraphics[width=\textwidth]{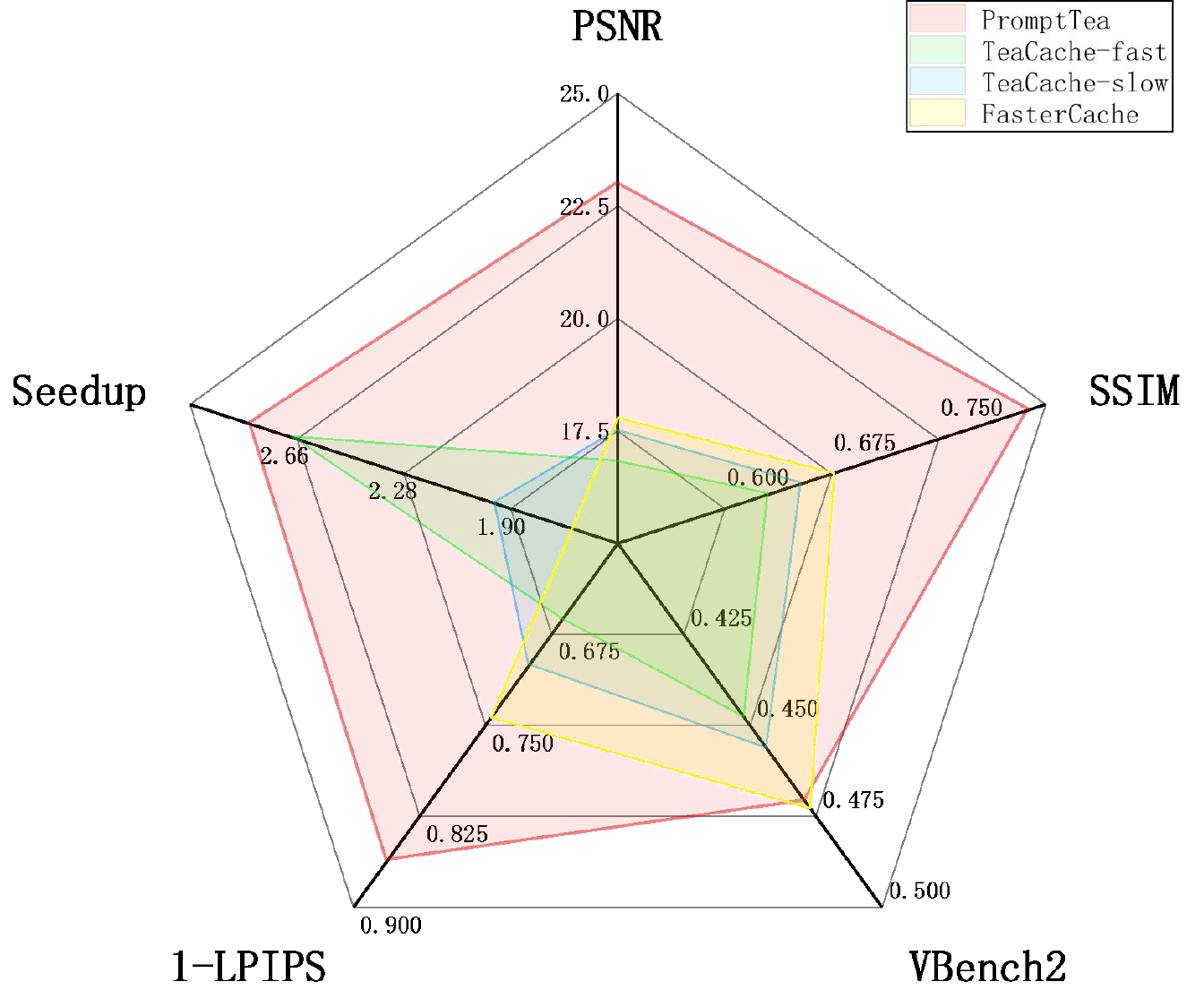}
        \caption{Numerical Scores}
        \label{fig.Numerical Scores}
    \end{subfigure}
\caption{Compares the performance of PromptTea and TeaCache. (a) Visual Quality: For the baselines of CogVideoX1.5, HunyuanVideo, and Wan2.1, the visual results of our PromptTea method are closer to the baselines compared to those of the TeaCache method, indicating better visual quality of our PromptTea. (b) Numerical Scores: The radar chart shows that PromptTea has better performance in metrics such as PSNR, SSIM, LPIPS, VBench2, and speedup. \textbf{Note that the details can be observed by magnifying the images. The same applies to the subsequent figures.}}
\label{fig.comparsion}
\end{figure*}

\begin{abstract}
Despite recent progress in video generation, inference speed remains a major bottleneck. A common acceleration strategy involves reusing model outputs via caching mechanisms at fixed intervals. However, we find that such fixed-frequency reuse significantly degrades quality in complex scenes, while manually tuning reuse thresholds is inefficient and lacks robustness. To address this, we propose Prompt-Complexity-Aware (PCA) caching, a method that automatically adjusts reuse thresholds based on scene complexity estimated directly from the input prompt. By incorporating prompt-derived semantic cues, PCA enables more adaptive and informed reuse decisions than conventional caching methods.
We also revisit the assumptions behind TeaCache and identify a key limitation: it suffers from poor input-output relationship modeling due to an oversimplified prior. To overcome this, we decouple the noisy input, enhance the contribution of meaningful textual information, and improve the model's predictive accuracy through multivariate polynomial feature expansion.
To further reduce computational cost, we replace the static CFGCache with DynCFGCache, a dynamic mechanism that selectively reuses classifier-free guidance (CFG) outputs based on estimated output variations. This allows for more flexible reuse without compromising output quality.
Extensive experiments demonstrate that our approach achieves significant acceleration—for example, 2.79× speedup on the Wan2.1 model—while maintaining high visual fidelity across a range of scenes.
\end{abstract}


\begin{links}
    \link{Code}{https://github.com/zishen-ucap/PromptTea.git}
\end{links}

\section{Introduction}

Video generation, particularly using diffusion models \cite{diffusion1, diffusion2, diffusion3, diffusion4}, has become a dominant paradigm in recent advancements. While the shift from traditional U-Net architectures \cite{unet1, unet2, unet3} to more complex DIT-based architectures \cite{dit} has significantly improved the quality of generated videos, it has also resulted in a dramatic increase in computational cost and inference latency. For example, state-of-the-art models such as Wan2.1 \cite{wan2025}, HunyuanVideo \cite{kong2024hunyuanvideo}, and CogVideoX1.5 \cite{yang2024cogvideox} require up to 50 inference steps per video, each step introducing considerable computational complexity. Given the increasing demand for longer video sequences and higher resolutions, this growing computational burden has become a critical bottleneck for efficient video generation.

To tackle this issue, several acceleration techniques have been proposed, such as model distillation \cite{distillation1, distillation12, distillation13} and quantization \cite{quantization}, but these methods often necessitate additional training, further increasing computational overhead. An alternative and increasingly popular strategy is the use of caching mechanisms, which allow for the reuse of intermediate denoising features without the need for retraining, thus reducing redundant computations. Early caching strategies, such as uniform caching \cite{pab, deltadit}, assumed that outputs at consecutive timesteps are similar. However, this assumption breaks down in models like CogVideoX1.5, leading to unnecessary computational redundancy during reuse.

In contrast, dynamic caching strategies \cite{teacache} selectively reuse cache across timesteps, aiming to minimize computational redundancy. While these methods are more flexible, they still face two key limitations: (1) TeaCache \cite{teacache} uses TEMNI (timestep embedding modulated noisy input) as an input to estimate relative output differences. However, we find that the noisy input has minimal contribution, often diluting the important textual semantics and skewing the estimated relationships. (2) Existing dynamic caching methods often perform inconsistently across scenes of varying complexity, experiencing greater quality degradation in more complex scenes and requiring time-consuming, non-robust threshold adjustments.

These shortcomings highlight the need for a more refined approach to video caching. Specifically, the assumptions behind TeaCache—such as the strong correlation between timestep embedding and TEMNI with model output—often lead to cumulative errors in iterative outputs. While FasterCache’s \cite{fastercache} CFGCache mechanism has leveraged classifier-free guidance (CFG) to accelerate inference, it, like earlier uniform caching methods, lacks the flexibility needed to adapt to variations in output differences between consecutive timesteps, ultimately limiting cache utilization.

To address the challenges in current caching mechanisms for video generation, we propose the following contributions:

\begin{itemize}
    \item \textbf{Prompt-Complexity-Aware (PCA) Caching}: We introduce a novel method that estimates scene complexity directly from textual prompts using cosine similarity of text embeddings. This enables automatic, adaptive thresholding for cache reuse while eliminating the interference of noisy inputs and better leveraging prompt semantics.
    \item  \textbf{Input-Output Relationship Refinement}: We replace the noisy TEMNI representation with cleaner inputs—timestep embeddings and scalar timestep values—and apply multivariate polynomial feature expansion to improve the accuracy of output similarity prediction.
    \item \textbf{Dynamic CFGCache}: We upgrade the static CFGCache into a dynamic mechanism that adaptively reuses CFG outputs by estimating output differences between consecutive timesteps, improving cache utilization and reducing redundancy.
\end{itemize}

Experiments show our method achieves 2.79× speedup on Wan2.1 with minimal quality loss (PSNR=23.0 dB), outperforming SOTA caching mechanisms, as shown in \text{Fig. \ref{fig.comparsion}}.

\section{Related work}
\subsection{Diffusion Model}

Recent advancements in video generation have predominantly relied on diffusion models \cite{diffusion1, diffusion2}, which have become a leading paradigm in generative tasks. Early video generation methods mainly used U-Net architectures \cite{unet1, unet2, unet3}, but their scalability limitations became apparent when generating long-duration, high-resolution content. The Diffusion Transformer (DiT) \cite{dit} was introduced to address these challenges by leveraging larger model capacities, significantly improving video quality. DiT’s attention mechanisms have evolved from 2+1D spatiotemporal attention \cite{opensora, opensoraplan} to full 3D attention \cite{yang2024cogvideox}, further enhancing model performance. However, these improvements have come with an increase in computational complexity, which has led to significant bottlenecks in inference speed.

\subsection{Diffusion Model Acceleration}
Acceleration Techniques for Diffusion Models
To mitigate the computational cost of diffusion models, various acceleration techniques have been proposed, categorized into training-dependent and training-free methods. Training-dependent methods, such as model distillation \cite{distillation1, distillation12} and quantization \cite{quantization}, can improve inference speed but introduce extra training overhead, making them less suitable for some applications.

Training-free methods, such as distributed inference \cite{li2024distrifusion}, parallelize the inference process across multiple computing nodes, reducing overall latency. However, this method requires substantial additional GPU memory for data replication and inter-node communication, making it impractical for certain real-time scenarios. Other training-free methods, such as caching mechanisms, have gained popularity due to their plug-and-play nature. Early caching mechanisms, such as DeepCache \cite{ma2024deepcache} and FasterDiffusion \cite{fasterdiffusion}, primarily targeted U-Net architectures by reusing features from intermediate layers to reduce computational redundancy. However, with the rise of DiT models, caching mechanisms have evolved. $\delta$-DiT \cite{deltadit} improves efficiency by selectively reusing partial blocks from consecutive timesteps, while PAB \cite{pab} introduces proportional and selective uniform caching to improve computational efficiency in DiT models. Despite these advancements, uniform caching strategies still fail to account for feature changes across different stages of the denoising process, leading to suboptimal cache utilization.

To overcome this, TeaCache \cite{teacache} introduced a dynamic caching strategy that adaptively selects which cache to reuse based on estimated output differences. However, the TEMNI input used in TeaCache dilutes the role of textual information, leading to degraded output quality. Moreover, TeaCache’s reliance on manual threshold adjustment introduces inefficiency, as the method struggles to adapt to varying scene complexities. FasterCache \cite{fastercache} introduces CFGCache, a mechanism that reuses cache based on CFG to accelerate inference. However, like early uniform caching methods, FasterCache still relies on a fixed caching strategy and does not dynamically adjust which timesteps' caches can be reused, limiting cache utilization efficiency.

\section{Methodology}
\subsection{Preliminaries}
\subsubsection{Denoising Diffusion Models}
Denoising diffusion models are a class of generative models that generate samples by iteratively denoising. The basic idea is to start with random noise and refine it step - by - step until it approximates a sample from the target distribution.During the forward diffusion process, Gaussian noise is added incrementally over a sequence of T steps to a data point $x_0$ sampled from the real - world distribution $q(x)$. This can be mathematically formulated as:
\begin{equation}
    x_t = \sqrt{\alpha_t}x_{t-1} + \sqrt{1-\alpha_t}z_t, \quad t=1,\ldots,T
    \label{eq:add noise}
\end{equation}

In this equation, $\alpha_t\in[0, 1]$controls the level of noise introduced at each step, and $z_t\sim \mathcal{N}(0, I)$ represents Gaussian noise. As$t$ increases, $x_t$ becomes increasingly noisy, and when $t = T$, it approaches a normal distribution $\mathcal{N}(0, I)$.

The reverse diffusion process aims to recover the original data from the noisy version. It is described by the following equation:
\begin{equation}
    p_\theta(x_{t-1} \mid x_t) = N(x_{t-1}; \mu_\theta(x_t, t), \Sigma_\theta(x_t, t))
    \label{eq:denoise}
\end{equation}

Here, $\mu_{\theta}$ and $\Sigma_{\theta}$ are learned parameters that define the mean and covariance.

\subsubsection{TeaCache}
\cite{teacache} proposed that in DiT models at discrete timesteps, there is a strong correlation between the inputs (e.g., time embedding and TEMNI) and outputs of the model at consecutive timesteps. Based on this correlation, the output difference can be estimated according to the input difference, and the decision of whether to reuse the cache can be made by the accumulated estimated output difference, which can be mathematically expressed as follows:

Let $f$ be the relationship formula of input - output, $\theta_{T_{ea}}$ be the learnable parameter in this formula, $F$ be TEMNI, $t$ be the timestep, $L1_{rel}(F,t)$ represent the relative difference of TEMNI at timestep $t$ and $t + 1$, and $\delta_{T_{ea}}$ be the caching threshold of the TeaCache algorithm, which is set manually. The mathematical relationship is:
\begin{equation}
    \sum_{t = t_a}^{t_b - 1} f\left(L1_{rel}(F,t),\theta_{Tea}\right) \leq \delta_{Tea} \leq \sum_{t = t_a}^{t_b} f\left(L1_{rel}(F,t),\theta_{T_{ea}}\right)
    \label{eq.teacache ori}
\end{equation}

This formula can be interpreted as follows: at timestep $t_a$, the model output is calculated and cached. As the timestep progresses, the accumulated estimated output difference is calculated. If the accumulated estimated output difference satisfies the left-hand side of the inequality (i.e., less than or equal to $\delta_{T_{ea}}$), it means that the output difference between the current timestep $t$ ($t_a < t < t_b$) and the timestep $t_a$ is small, and the model output at timestep $t_a$ can be reused; otherwise, the model output needs to be recalculated through dit, and the cache is updated to the model output of the current timestep, while the accumulated estimated output difference is reset to zero.

The relative difference $L1_{rel}(F,t)$ can be expressed as:
\begin{equation}
    L1_{rel}(F,t) = \frac{\left\| F_t - F_{t+1} \right\|_1}{\left\| F_{t+1} \right\|_1}
    \label{eq.l1}
\end{equation}

where $F_t$ represents TEMNI at timestep $t$.

\subsection{Analysis} \label{analysis}

\begin{figure}[htbp]
\centering
    \begin{subfigure}[b]{0.15\textwidth}
        \includegraphics[width=\textwidth]{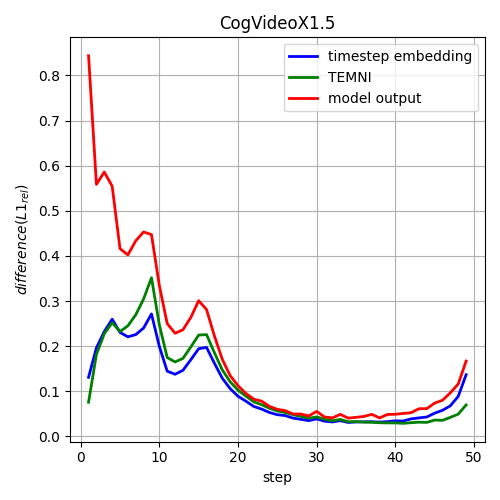}
        \caption{CogVideoX1.5}
        \label{fig.cog diff}
    \end{subfigure}
    \hfill
    \begin{subfigure}[b]{0.15\textwidth}
        \includegraphics[width=\textwidth]{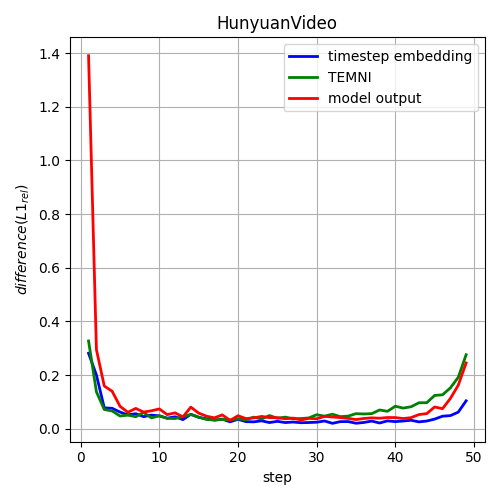}
        \caption{HunyuanVideo}
        \label{fig.hun diff}
    \end{subfigure}
    \hfill
    \begin{subfigure}[b]{0.15\textwidth}
        \includegraphics[width=\textwidth]{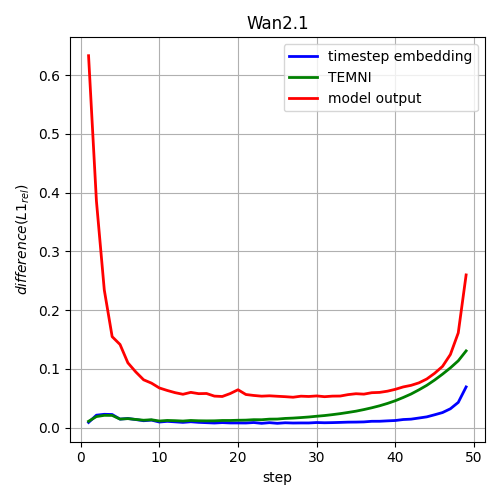}
        \caption{Wan2.1}
        \label{fig.wan diff}
    \end{subfigure}
\caption{Comparison of correlations among timpstep embedding, TEMNI, and output. }
\label{fig.all diff}
\end{figure}

\begin{figure}[htbp]
    \centering
    \includegraphics[width=0.9\linewidth]{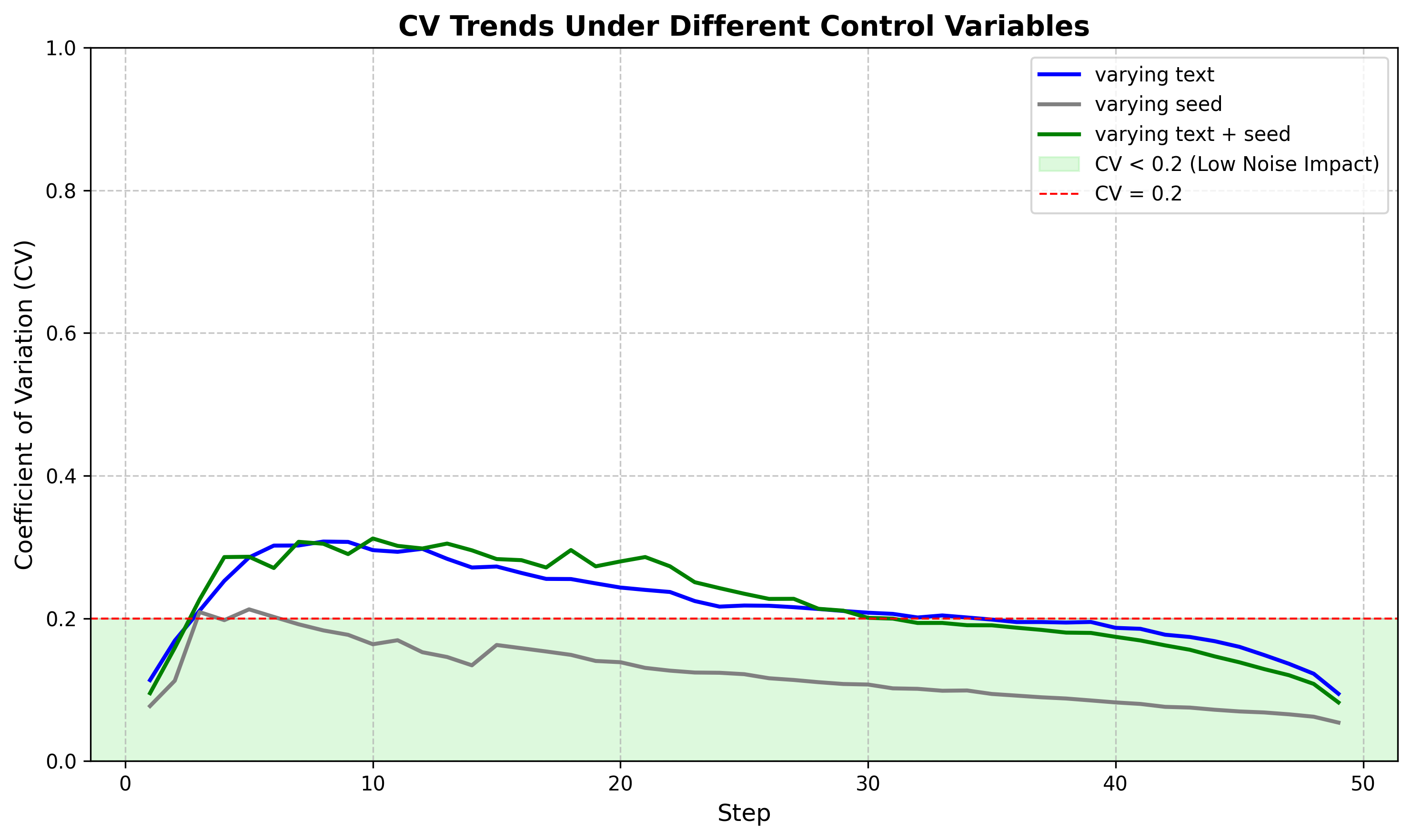}
    \caption{CV trends under controlled text and seed variables, showing text variation impacts output difference significantly while seed variation has minimal effect, indicating noisy input (seed) has little influence on output difference.}
    \label{fig:vary}
\end{figure}

\subsubsection{Input-Output Correlation Analysis}
Studies on 3D full-attention DiT models have shown that strong input-output correlations do not universally hold across all models. As shown in \text{Fig. \ref{fig.all diff}}, CogVideoX1.5 and Wan2.1 exhibit weak correlations in the first half of timesteps. While TeaCache typically uses TEMNI (incorporating text, noisy input, and timestep information) as input, claiming a stronger correlation with outputs than timestep embeddings alone, experiments on CogVideoX1.5, Hunyuanvideo, and Wan2.1 show that timestep embeddings correlate more strongly with outputs (\text{Fig. \ref{fig.all diff}}).

Further analysis of the impact of text and noisy input on output differences using the control variable method: with 50 different seeds and 50 different prompts as samples, when the text is fixed and the seed is varied, the coefficient of variation (the ratio of standard deviation to mean) for most timesteps is below 0.2, indicating that random noise has a limited impact on output differences; when the seed is fixed and the text is varied, or both the text and seed are varied, the coefficient of variation is generally above 0.2, and the results of the two groups are close (\text{Fig. \ref{fig:vary}}). This leads to the conclusion that noisy input has a minimal effect on output differences, whereas the text is the dominant factor. Therefore, decoupling noisy input and deciding whether to reuse the cache based solely on text and timestep information becomes a key issue.

\subsubsection{Video Caching Threshold Specificity}
\begin{figure}[htbp]
    \centering
    \includegraphics[width=0.9\linewidth]{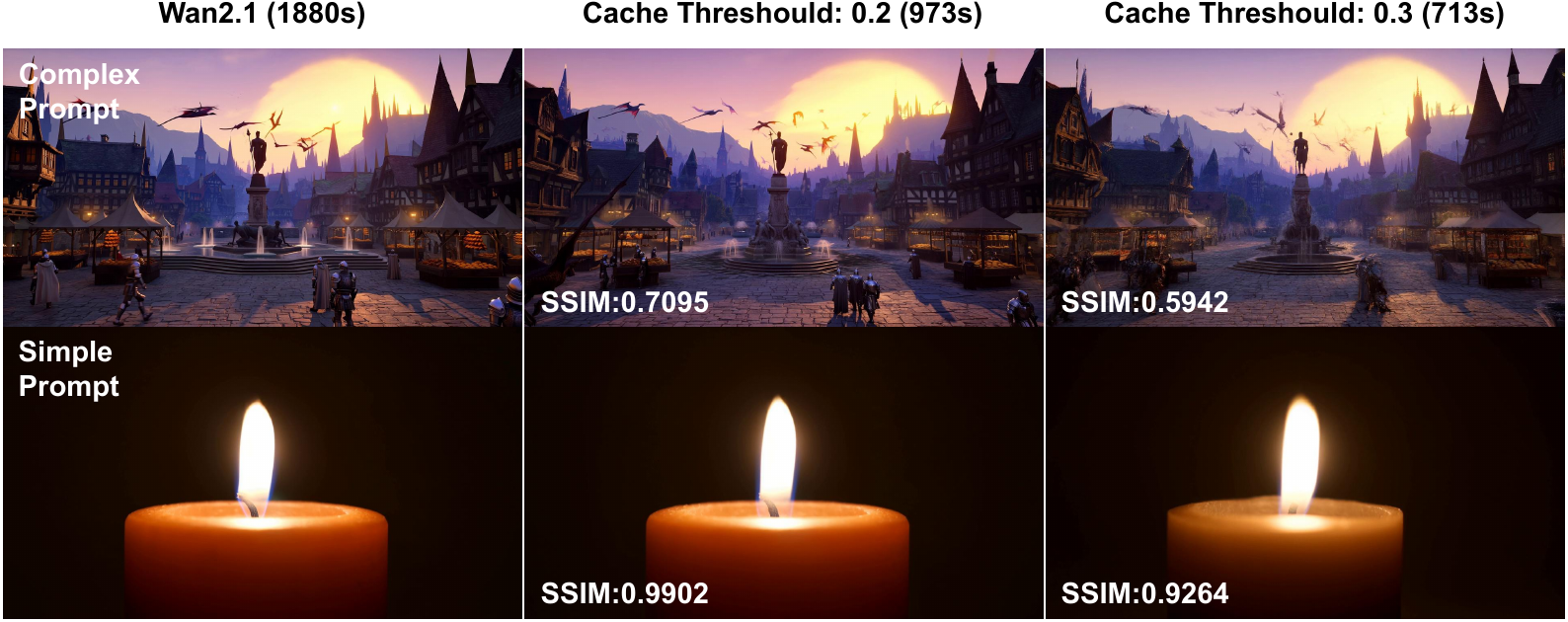}
    \caption{In TeaCache, different caching thresholds lead to varying performance of prompts with different complexity levels.}
    \label{fig.different threshould}
\end{figure}

\cite{adacache} proposes that the minimum number of timesteps required for generating videos varies across different cases. The caching mechanism can be viewed as a special case of timestep reduction, where the optimal caching threshold differs for each video in dynamic caching. As shown in \text{Fig. \ref{fig.different threshould}}, setting the caching threshold to 0.3 in simple prompt scenarios does not significantly affect the visual quality, while in more complex prompt scenarios, the threshold needs to be lowered to 0.2 in order to maintain visual fidelity. This reveals a strong correlation between the complexity of video generation scenarios (determined by the text prompts) and the caching threshold: simpler scenarios can tolerate higher thresholds to improve speed, while more complex ones require lower thresholds to ensure quality. However, existing methods (such as TeaCache) fail to explicitly model text complexity, making it impossible to adaptively adjust the threshold. Therefore, dynamically determining the optimal threshold for each video based on text information presents a significant challenge in balancing quality and efficiency.

\subsection{PromptTea}
Based on the findings in the ‘Analysis’ section, we propose the PCA-TeaCache framework, which quantifies scenario complexity through text embeddings to achieve adaptive adjustment of caching thresholds. The framework has two main functions:
\begin{itemize}
    \item \textbf{Eliminate noisy input interference}: Use timestep embeddings instead of noisy TEMNI as inputs to avoid the impact of noisy input on output difference estimation; and supplement text information for cache reuse in the form of thresholds.
    \item  \textbf{Dynamically adapt thresholds}: Assign optimal caching thresholds to video generation scenarios of different complexities based on the clustering characteristics of text embeddings.
\end{itemize} 

\subsubsection{Prompt Complexity Aware}
\begin{figure*}[htbp]
    \centering
    \includegraphics[width=0.75\linewidth]{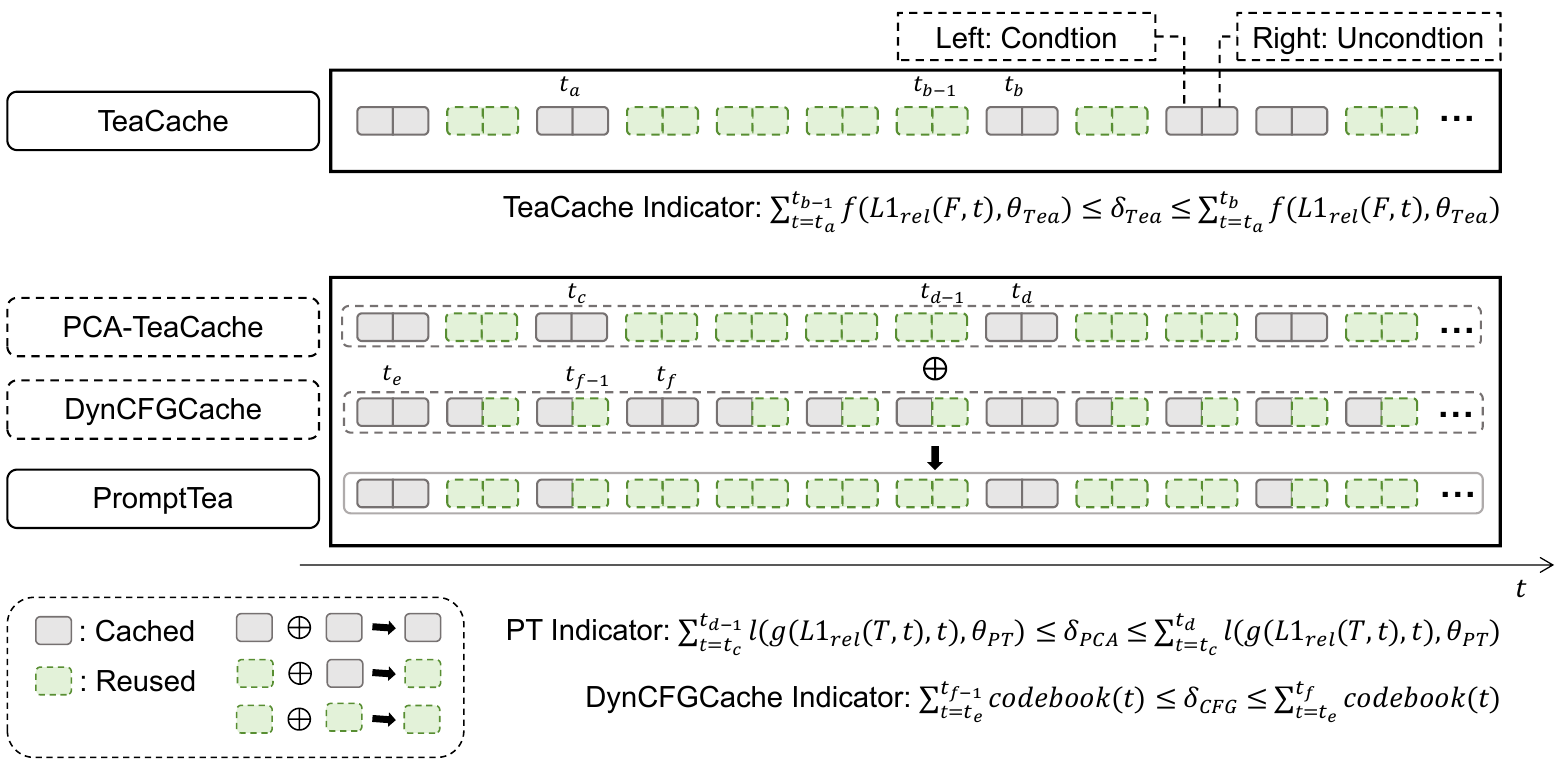}
    \caption{PromptTea consists of PCA-TeaCache (an improved TeaCache eliminating noisy input and adapting thresholds via text embeddings) and DynCFGCache (using dynamic difference accumulation for caching). Their combination enhances efficiency and accuracy. Left/right sub-squares denote conditional/unconditional CFG feature calculations; gray squares indicate computing/caching, green ones reuse. PT Indicator and DynCFGCache Indicator refer to \text{Eq. \ref{eq.PCA caching indicator}} and \text{Eq. \ref{eq.dyncfgcache}}.}
    \label{fig.toc}
\end{figure*}
\begin{figure}[htbp]
    \centering
    \includegraphics[width=1\linewidth]{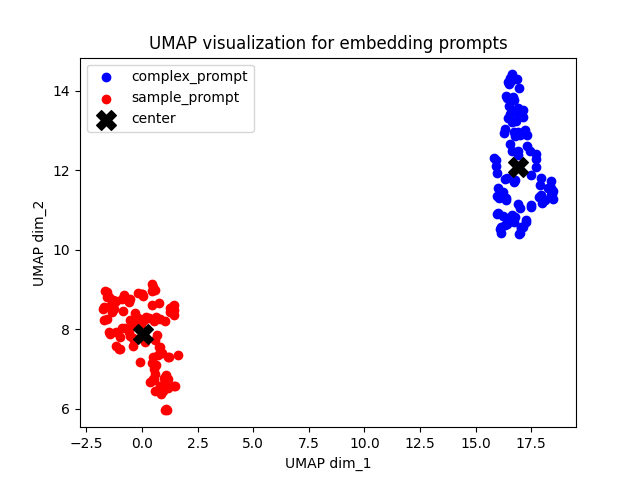}
    \caption{After UMAP dimensionality reduction, the text embeddings of complex prompts and simple prompts form distinct clusters, indicating that simple prompts are more similar to each other, and complex prompts are more similar to each other.}
    \label{fig.prompt_embedding_visualizationl}
\end{figure}
\begin{figure}[htbp]
    \centering
    \includegraphics[width=0.9\linewidth]{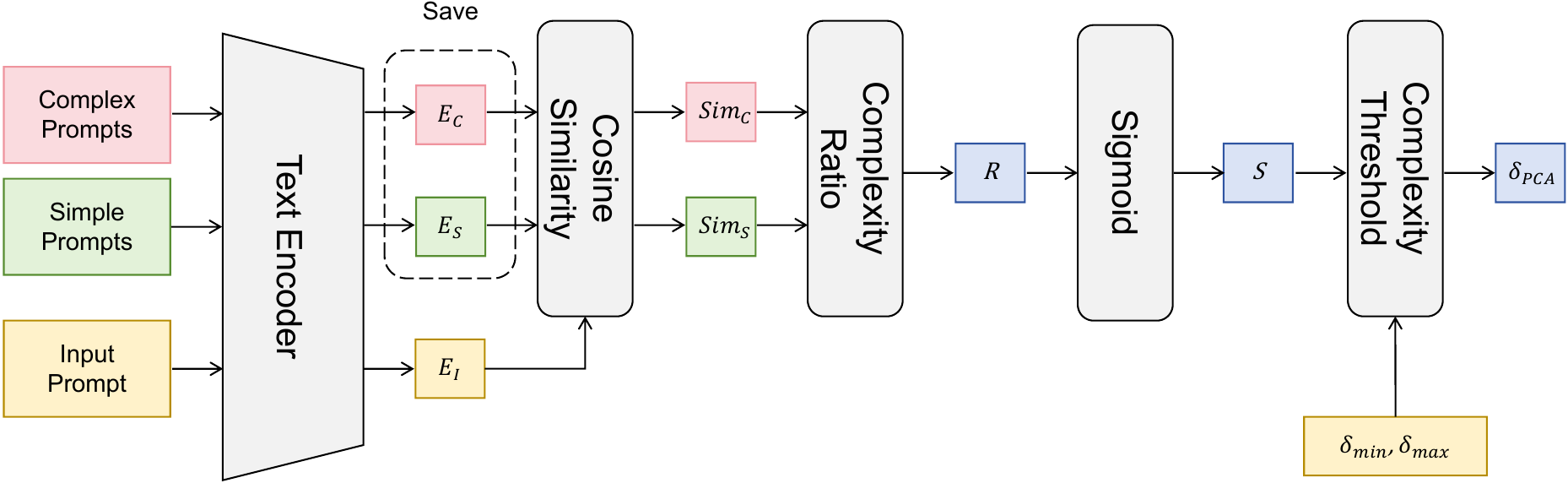}
    \caption{Flow chart of Prompt Complexity Aware (PCA)}
    \label{fig.PCAl}
\end{figure}

To eliminate interference from noisy input when estimating output differences, we use timestep embeddings instead of TEMNI as inputs. However, text information must be incorporated to improve estimation accuracy. We generated 100 complex prompts (defined as descriptions with multiple subjects and rich dynamic details) and 100 simple prompts (involving single subjects and low-dynamic scenarios) using DeepSeek \cite{deepseekai2024deepseekv3technicalreport}, and extracted their embedding vectors via the text encoder of the video generation model. Experimental results show that the average embedding distance among complex prompts is 1.6955, while that among simple prompts is 1.1758; the cross-average distance between the two types of prompts is 2.3224. After dimensionality reduction using UMAP \cite{mcinnes2018umap}, the silhouette coefficient of the two sets of embedding vectors reaches 0.9041 (see \text{Fig. \ref{fig.prompt_embedding_visualizationl}})), indicating that the text embeddings of simple and complex prompts form distinct clusters.

Based on the above findings, we propose to evaluate scenario complexity by comparing the cosine similarity between the input prompt and pre-saved complex/simple prompt sets' text embeddings (as shown in \text{Fig. \ref{fig.PCAl}}): first, pre-save the text embeddings $E_C$ of the complex prompt set and $E_S$ of the simple prompt set, and calculate the cosine similarities $Sim_C$ and $Sim_S$ between the input prompt's text embedding $E_I$ and $E_C$, $E_S$.

Based on the above findings, we propose to evaluate scenario complexity by comparing the cosine similarity between the input prompt’s text embedding and the pre-saved text embeddings of complex/simple prompt sets (as shown in \text{Fig. \ref{fig.PCAl}}). The process is as follows: first, pre-save the text embeddings $E_{C}$ of the complex prompt set and $E_{S}$ of the simple prompt set; then, calculate the cosine similarities $Sim_{C}$ and $Sim_{S}$ between the input prompt’s text embedding $E_{I}$ and $E_{C}$, $E_{S}$ respectively.

The complexity coefficient $R$ is defined as:
\begin{equation}
    R = \frac{Sim_C}{Sim_C + Sim_S + \varepsilon}
    \label{eq.complexity coefficient}
\end{equation}

where $\varepsilon = 10^{-6}$ avoids division by zero, and a larger $R$ indicates a more complex scenario. The complexity coefficient $R$ is then stretched to the interval boundary using the Sigmoid function $S$:
\begin{equation}
    S = \frac{1}{1 + e^{-k (R - 0.5)}}
    \label{eq.sigmoid}
\end{equation}

where $k$ controls the stretching amplitude. Finally, the adaptive threshold is obtained by weighted fusion of the upper and lower threshold limits $\delta_{\text{max}}$ and $\delta_{\text{min}}$:
\begin{equation}
    \delta_{PCA}=S\cdot\delta_{\text{max}}+(1-S)\cdot\delta_{\text{min}}
    \label{eq.PCA threshould}
\end{equation}

This threshold achieves dynamic adaptation of "low threshold for complex scenarios and high threshold for simple scenarios", echoing the "threshold specificity" requirement proposed in the "Video Caching Threshold Specificity" section.

\subsubsection{Input-Output Relationship Reconstruction}
Aiming at the problem in CogVideoX1.5 and Wan2.1 where "input differences are similar but output differences are significant" (such as timesteps 1 and 19 in Wan2.1, see Fig. \ref{fig.all diff}), which leads to large fitting errors in TeaCache (see Appendix), we extend input features through multivariate polynomial mapping—i.e., by introducing positional information to inform the fitting model of its current sequence position—to enhance the modeling capability for complex relationships.

The relative difference of timestep embeddings $x=L1_{\text{rel}}(\mathbf{T}, t)$ and the timestep $t$ are combined into 4th-order polynomial features:
\begin{equation}
   X = g(x, t) = [1\ x\ t\ x^2\ t^2\ xt\ x^3\ t^3\ x^2t\ xt^2\ x^4\ t^4]^{\top}
   \label{eq.T and t}
\end{equation}

Define $y$ as the model output difference (i.e., the relative difference value of adjacent timestep outputs). We use $100 \times 50$ groups of $(X,y)$ data, with each group including 100 prompts, each prompt having 50 timesteps of input features $X$ and corresponding output difference $y$. Then, the model $l$ is fitted using \texttt{sklearn.linear\_model.LinearRegression}. Feature expansion effectively improves the accuracy of linear relationship modeling, as evidenced by the fact that the fitting curve of our method is closer to the real $y$ value compared with TeaCache. The detailed experiments regarding this are placed in the Appendix.

The final caching indicator is formulated as:
\begin{equation}
    \sum_{t = t_c}^{t_d - 1} l\left(X,\theta_{PT}\right) \leq \delta_{PCA} < \sum_{t = t_c}^{t_d} l\left(X,\theta_{PT}\right)
    \label{eq.PCA caching indicator}
\end{equation}

Here, $\theta_{PT}$ denotes the parameters of the linear model fitted to $X$ and $y$, and the $\delta_{PCA}$ used here is consistent with the $\delta_{PCA}$ derived in \text{Eq. \ref{eq.PCA threshould}}. The decision logic aligns with that in \text{Eq. \ref{eq.teacache ori}}, but feature engineering significantly improves decision accuracy in complex scenarios. We refer to the caching mechanism that combines PCA with this reconstructed relationship as PCA-TeaCache.

\subsubsection{DynCFGCache}
Building on FasterCache’s CFGCache \cite{fastercache}, the DynCFGCache dynamic caching mechanism precomputes and averages a relative difference codebook for each timestep by collecting conditional and unconditional feature data from 100 prompts. As shown in Fig. 8, during inference, the codebook value for the current timestep is queried, accumulated, and compared with the threshold $delta_{CFG}$ to determine whether to reuse the cache. The final caching indicator is formulated as:

\begin{equation}
    \sum_{t = t_e}^{t_f - 1} codebook\left(t\right) \leq \delta_{CFG} < \sum_{t = t_e}^{t_f} codebook\left(t\right)
    \label{eq.dyncfgcache}
\end{equation}

This decision logic aligns with that in \text{Eq. \ref{eq.teacache ori}}. When the reuse condition is met, the unconditional feature output directly reuses the conditional feature output. Detailed differences between DynCFGCache and CFGCache are provided in the Appendix.

Additionally, DynCFGCache forms a collaborative mechanism with PCA-TeaCache: when both mechanisms decide to reuse the cache, PCA-TeaCache is prioritized to enhance decision accuracy in complex scenarios.

\section{Experiment}
\begin{table*} [htbp]
\centering
\caption{Quantitative evaluation of inference efficiency and visual quality in video generation models.}
\begin{tabular}{c|c c c|c c c c} 
\toprule[2pt]
\multirow{2}{*}{Method} & \multicolumn{3}{c|}{Efficiency} & \multicolumn{4}{c}{Visual Quality} \\
\cline{2-8}
& FLOPs (P) $\downarrow$ & Speedup $\uparrow$ & Latency (s) $\downarrow$ & VBench2 $\uparrow$ & LPIPS $\downarrow$ & SSIM $\uparrow$ & PSNR $\uparrow$ \\
\midrule
\midrule
\multicolumn{8}{c}{CogVideoX1.5 (81 frames, 1360$\times$768)} \\
\midrule
\rowcolor{gray!30}
CogVideoX1.5 ($T = 50$) & 150.12 & 1$\times$ & 485.84 & 0.4633 & - & - & - \\

FasterCache & 93.92 & 1.58$\times$ & 307.64 & 0.4767 & 0.3639 & 0.6041 & 17.03 \\

TeaCache-slow & 111.67 & 1.32$\times$ & 367.56 & 0.4669 & \textbf{0.0214} & \textbf{0.9540} & \textbf{36.26} \\

TeaCache-fast & 99.83 & 1.47$\times$ & 329.58 & 0.4604 & 0.4712 & 0.5169 & 15.04 \\

PromptTea & \textbf{84.63} & \textbf{1.75$\times$} & \textbf{278.14} & \textbf{0.4963} & 0.0836 & 0.8878 & 27.78 \\

\midrule
\midrule
\multicolumn{8}{c}{HunyuanVideo (129 frames, 1280$\times$720)} \\
\midrule
\rowcolor{gray!30}
HunyuanVideo ($T = 50$) & 85.16 & 1$\times$ & 1825.67 & 0.4875 & - & - & - \\

TeaCache-slow & 52.80 & 1.61$\times$ & 1130.90 & 0.4125 & 0.1477 & 0.8083 & 24.02 \\

TeaCache-fast & 39.85 & 2.17$\times$ & 842.22 & 0.4318 & 0.1554 & 0.8011 & 23.68 \\

PromptTea & \textbf{33.12} & \textbf{2.56$\times$} & \textbf{689.37} & \textbf{0.4444} & \textbf{0.1468} & \textbf{0.8138} & \textbf{24.42} \\

\midrule
\midrule
\multicolumn{8}{c}{WAN2.1 (129 frames, 1280$\times$720)} \\
\midrule
\rowcolor{gray!30}
Wan2.1 ($T = 50$) & 181.62 & 1$\times$ & 1880.34 & 0.5235 & - & - & - \\

FasterCache & 112.85 & 1.66$\times$ & 1135.53 & \textbf{0.4727} & 0.2565 & 0.6518 & 17.79 \\

TeaCache-slow & 91.14 & 1.93$\times$ & 972.92 & 0.4565 & 0.2993 & 0.6290 & 17.50 \\

TeaCache-fast & 66.81 & 2.64$\times$ & 713.35 & 0.4480 & 0.3371 & 0.6057 & 16.81 \\

PromptTea & \textbf{62.50} & \textbf{2.79$\times$} & \textbf{674.46} & 0.4710 & \textbf{0.1380} & \textbf{0.7884} & \textbf{23.00} \\

\bottomrule[2pt]
\end{tabular}
\label{table.comparison}
\end{table*}

\begin{table*} [htb!]
\centering
\caption{Ablation experiments of PromptTea modules. The values in parentheses represent the threshold of the method. PromptTea is defined as TeaCache+ $R$+$D$+PCA. The numbers inside the parentheses of PromptTea(0.1, 0.23) represent that the upper and lower threshold limits $\delta_{\text{max}}$ and $\delta_{\text{min}}$ in the \text{Eq. \ref{eq.PCA threshould}}  are 0.1 and 0.23 respectively.}
\begin{tabular}{c|c c c|c c c c} 
\toprule[2pt]
\multirow{2}{*}{Method} & \multicolumn{3}{c|}{Efficiency} & \multicolumn{4}{c}{Visual Quality} \\
\cline{2-8}
& FLOPs (P) $\downarrow$ & Speedup $\uparrow$ & Latency (s) $\downarrow$ & VBench2 $\uparrow$ & LPIPS $\downarrow$ & SSIM $\uparrow$ & PSNR $\uparrow$ \\
\midrule

TeaCache(0.2) & 91.14 & 1.93$\times$ & 972.92 & 0.4565 & 0.2993 & 0.6290 & 17.50 \\

TeaCache(0.3) & 66.81 & 2.64$\times$ & 713.35 & 0.4480 & 0.3371 & 0.6057 & 16.81 \\

TeaCache(0.2)+$R$ & 65.81 & 2.67$\times$ & 705.08 & 0.4684 & 0.1670 & 0.7515 & 21.60 \\

TeaCache(0.3)+$R$ & 51.33 & 3.39$\times$ & 554.24 & 0.4189 & 0.2285 & 0.6997 & 19.75 \\

TeaCache(0.2)+$R$+$D$ & 49.52 & 3.45$\times$ & 544.38 & \textbf{0.4729} & 0.1739 & 0.7523 & 21.61 \\

TeaCache(0.3)+$R$+$D$ & \textbf{44.09} & \textbf{3.85$\times$} & \textbf{487.88} & 0.3669 & 0.2312 & 0.7024 & 19.82 \\

PromptTea(0.1, 0.23) & 62.50 & 2.79$\times$ & 674.46 & 0.4710 & \textbf{0.1380} & \textbf{0.7884} & \textbf{23.00} \\

\bottomrule[2pt]
\end{tabular}
\label{table.ablation}
\end{table*}
\subsection{Implementation Details}
All experiments were conducted on Nvidia H100 80G GPUs with FlashAttention enabled by default. To ensure reproducibility, prompt extension was disabled, and other parameters followed the baselines’ official defaults. For key parameters: the stretching amplitude $k$ in \text{Eq. \ref{eq.sigmoid}} was set to 50 (CogVideoX1.5), 200 (HunyuanVideo), and 50 (Wan2.1); the PCA weighted fusion threshold bounds $(\delta_{\text{min}}, \delta_{\text{max}})$ in \text{Eq. \ref{eq.PCA threshould}} were (0.2, 0.3), (0.1, 0.15), and (0.1, 0.23) for the three models respectively; $\delta_{CFG}$ in \text{Eq. \ref{eq.dyncfgcache}} was 0.02 for CogVideoX1.5 and Wan2.1. Notably, HunyuanVideo conflicts with CFGCache due to incompatibility with its text-guidance distillation technique \cite{distillation1}, so it does not incorporate the DynCFGCache module. Correspondingly, in comparative experiments, the HunyuanVideo baseline does not include comparisons with FasterCache.

\subsection{Metrics}
This work evaluates video generation acceleration across two dimensions: inference efficiency (FLOPs, latency) and visual quality (VBench2, LPIPS, PSNR, SSIM). VBench2 \cite{zheng2025vbench2} is a human-aligned video benchmark; LPIPS \cite{lpips} measures perceptual similarity via pre-trained features; PSNR assesses pixel-level fidelity; SSIM evaluates structural similarity across luminance, contrast, and texture.

Due to high computational costs (e.g., Wan2.1 takes 31 minutes per video) and the impracticability of full VBench2 testing (63+ days for 1013 prompts × 3 seeds), we use random sampling: 5 prompts per dimension from VBench2’s 16 evaluation axes, balancing coverage and efficiency.

\subsection{Comparison Experiment}
In the comparison experiments, we set CogVideoX1.5 \cite{yang2024cogvideox}, HunyuanVideo \cite{kong2024hunyuanvideo}, and Wan2.1 \cite{wan2025} as the test baselines, and compare our PromptTea algorithm with three caching mechanism algorithms, TeaCache-slow, TeaCache-fast \cite{teacache} and FasterCache \cite{fastercache}.

\text{Tab. \ref{table.comparison}} conducts a quantitative evaluation of algorithm efficiency and visual quality using the VBench2 benchmark. Among them, PromptTea shows excellent efficiency and visual quality under different baseline models. When using CogVideoX1.5 as the baseline, although the visual quality of PromptTea is slightly inferior to that of TeaCache-slow, its efficiency advantage is significant, achieving a 1.75$\times$ speedup, higher than the 1.32$\times$ speedup of TeaCache - slow and the 1.47$\times$ speedup of TeaCache-fast. Moreover, the visual quality of PromptTea is better than that of TeaCache-fast. When using HunyuanVideo as the baseline, PromptTea achieves a 2.56$\times$ speedup, which is better than the 2.17$\times$ speedup of TeaCache-fast, and its visual quality surpasses that of TeaCache-slow. Similarly, when using Wan2.1 as the baseline, PromptTea reaches a 2.79$\times$ speedup, exceeding the 2.64$\times$ speedup of TeaCache-fast, and its visual quality is also better than that of TeaCache-slow.

\text{Fig. \ref{fig.Visual Quality}} shows the videos generated by PromptTea, TeaCache, and the baseline. The comparison results indicate that PromptTea outperforms TeaCache in both inference efficiency and visual quality. For more visual comparisons, please refer to the Appendix.

\subsection{Ablation Study}
We conduct ablation studies on three modules: Input-Output Relationship Reconstruction (denoted as $R$), DynCFGCache (denoted as $D$), and PCA, to verify the effectiveness of each module. Wan2.1 is used as the baseline for this ablation study.

\subsubsection{Ablation on Input-Output Relationship Reconstruction}
As shown in \text{Tab. \ref{table.ablation}}, comparing TeaCache(0.2) with TeaCache(0.2)+R after Input - Output Relationship Reconstruction, the latter has higher inference efficiency and better visual quality. Comparing TeaCache(0.3) with TeaCache(0.3)+R, although the VBench2 metric decreases, other visual metrics improve significantly, and the speedup increases from $2.64\times$ to $3.39\times$. These two groups of comparisons demonstrate that the Input-Output Relationship Reconstruction accurately corrects the input - output relationship, enabling the model to generate better visual quality at a faster speed.

\subsubsection{Ablation on DynCFGCache}
When DynCFGCache is introduced to TeaCache+$R$, it becomes PromptTea without the PCA module. By comparing TeaCache(0.2)+$R$ with TeaCache(0.2)+$R$+$D$, we observe that the visual quality slightly improves rather than decreases, and the speedup increases from $2.67\times$ to $3.45\times$. By comparing TeaCache(0.3)+$R$ with TeaCache(0.3)+$R$+$D$, we find that although VBench2 decreases, other visual metrics improve, and the speedup increases from $3.39\times$ to $3.85\times$. These two groups of comparisons indicate that DynCFGCache can improve the model's inference speed with negligible degradation in visual quality.

\subsubsection{Ablation on PCA}
We found that when the threshold is set to 0.3, the VBench2 score decreases significantly. Considering the balance between visual quality and inference speed, we set the threshold boundaries to (0.1, 0.23). Then, by comparing PromptTea(0.1, 0.23) with TeaCache(0.2)+$R$+$D$ and TeaCache(0.3)+$R$+$D$, we find that PromptTea(0.1, 0.23) achieves a balance in visual quality at the cost of a slight decrease in inference speed.

\section{Conclusion}
In this research, we propose PromptTea, a method dynamically choosing caching thresholds based on prompt complexity. For complex videos, low thresholds maintain quality; for simple ones, high thresholds ensure speed. By decoupling noisy input, text, and timestep, we corrected TeaCache's input-output relationship, as noisy input weakens text information. The proposed PCA adapts thresholds to video complexity for speed-quality balance, while DynCFGCache—upgraded from uniform CFGCache—reduces computational redundancy. Experiments show PromptTea achieves significant acceleration with minimal visual quality loss, validating its effectiveness in optimizing DiT-based diffusion models.

\newpage

\bigskip

\bibliography{aaai25}

\clearpage

\begin{figure*}[htbp]
\centering
\includegraphics[width=0.9\linewidth]{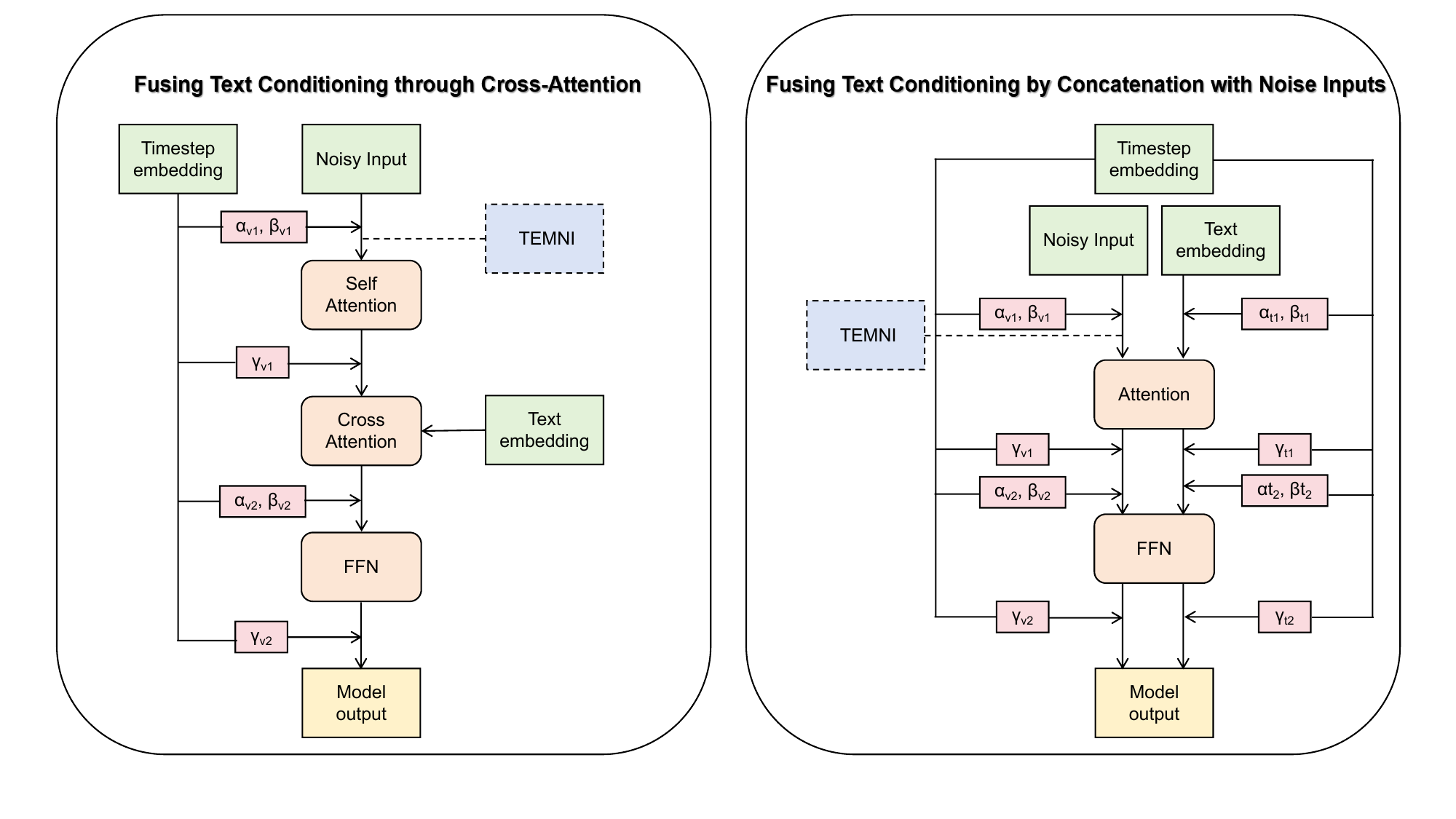} 
\caption{The locations of TEMNI in the two frameworks. Blue squares denote the positions of TEMNI modules}
\label{fig.TEMNI}
\end{figure*}

\section{Appendix}

\subsection{Timestep Embedding Modulated Noisy Input}
TEMNI (Timestep Embedding Modulated Noisy Input) serves as the input for TeaCache for two key reasons. First, it integrates the three core inputs of the DiT block: text embedding, timestep embedding, and noisy input. Second, prior research has demonstrated that TEMNI exhibits a strong correlation with output differences, making it a critical component for TeaCache to capture generative discrepancies.

The DiT model has developed into two mainstream frameworks: one that fuses text conditioning through cross-attention, and another that fuses text conditioning by concatenation with noise inputs. Correspondingly, the position of TEMNI differs between these two frameworks.

As illustrated in \text{Fig. \ref{fig.TEMNI}}, within the first framework (Fusing Text Conditioning through Cross-Attention), TEMNI is situated before the self-attention module. In the second framework (Fusing Text Conditioning by Concatenation with Noise Inputs), TEMNI is placed before the attention module—specifically, this position occurs prior to concatenation with text embedding.

\subsection{Prompt Can Reflect the Complexity of Generated Videos}
\begin{table}[htbp]
\centering
\caption{Comparison of PSNR and Average Optical Flow Amplitude for videos generated by Simple and Complex Prompts..}
\begin{tabular}{c|cc}
\toprule[2pt]
Prompt Type & PSNR & Average Optical Flow Amplitude \\
\midrule
Simple & 36.52 & 1.4551  \\
Complex & 32.92 & 2.9748  \\
\bottomrule[2pt]
\end{tabular}
\label{table.prompt type}
\end{table}

In diffusion - based video generation, prompts play a significant role in determining the characteristics of the generated videos. Here, we use the Wan2.1 model with 100 simple prompts and 100 complex prompts for video generation. To evaluate the motion amplitudes of the generated videos, we measure the PSNR(dB) between adjacent frames and the average optical flow amplitude(pixels/frame).

As shown in \text{Tab. \ref{table.prompt type}}, videos from simple prompts have a higher PSNR (36.52 dB) and lower average optical flow amplitude (1.4551 pixels/frame), indicating less motion. In contrast, videos from complex prompts have a lower PSNR (32.96 dB) and higher average optical flow amplitude (2.9748 pixels/frame), showing more motion. This demonstrates that complex prompts result in videos with larger motion amplitudes.

\subsection{More Details on Input-Output Relationship Reconstruction}
\begin{figure*}[htbp]
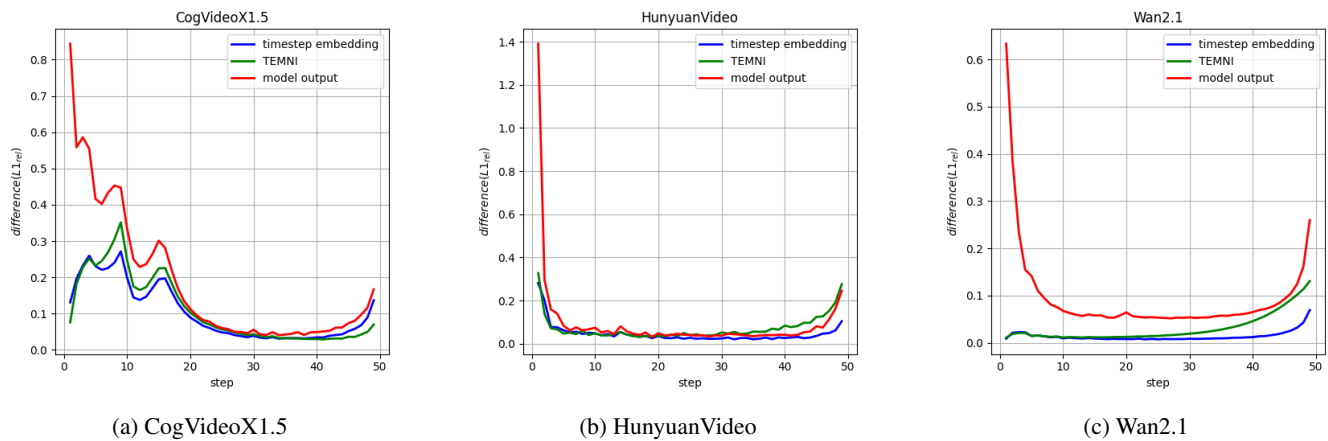

\centering
    \begin{subfigure}[b]{0.3\textwidth}
        \includegraphics[width=\textwidth]{img/cog-diff.png}
        \caption{CogVideoX1.5}
        \label{fig.cog diff2}
    \end{subfigure}
    \hfill
    \begin{subfigure}[b]{0.3\textwidth}
        \includegraphics[width=\textwidth]{img/hun-diff.png}
        \caption{HunyuanVideo}
        \label{fig.hun diff2}
    \end{subfigure}
    \hfill
    \begin{subfigure}[b]{0.3\textwidth}
        \includegraphics[width=\textwidth]{img/wan-diff.png}
        \caption{Wan2.1}
        \label{fig.wan diff2}
    \end{subfigure}
\caption{Comparison of correlations among timpstep embedding, TEMNI, and output. }
\label{fig.all diff2}
\end{figure*}

\begin{figure*}[htbp]
\centering
    \begin{subfigure}[b]{0.33\textwidth}
        \includegraphics[width=\textwidth]{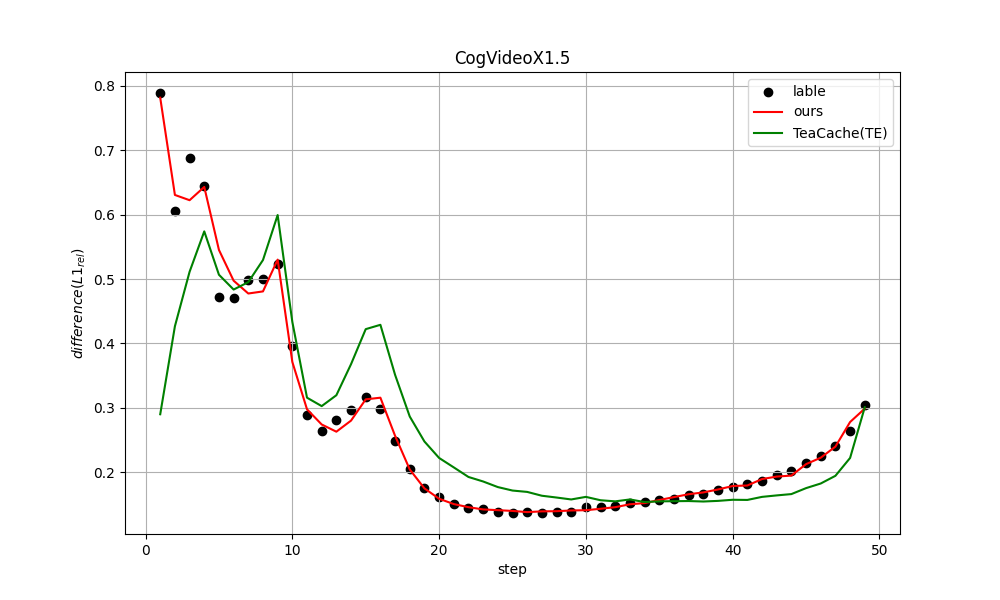}
        \caption{CogVideoX1.5}
        \label{fig.cog fit}
    \end{subfigure}
    \hfill
    \begin{subfigure}[b]{0.33\textwidth}
        \includegraphics[width=\textwidth]{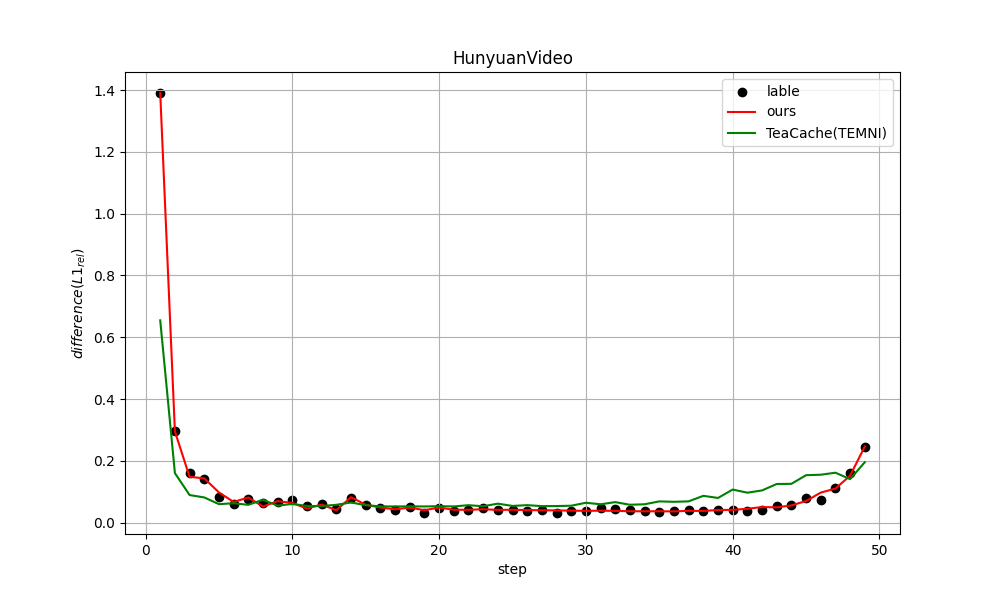}
        \caption{HunyuanVideo}
        \label{fig.hun fit}
    \end{subfigure}
    \hfill
    \begin{subfigure}[b]{0.33\textwidth}
        \includegraphics[width=\textwidth]{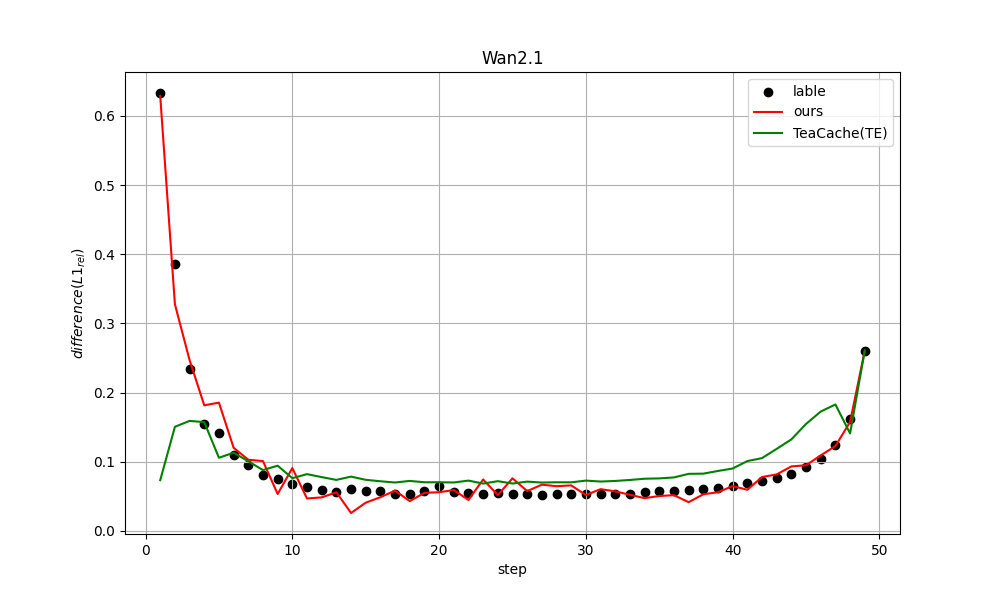}
        \caption{Wan2.1}
        \label{fig.wan fit}
    \end{subfigure}
\caption{Input-output relationship fitting results: The green curves denote our polynomial fitting, the red curves represent TeaCache's model, and black dots indicate ground-truth label values. Comparative analysis shows that our fitting demonstrates significantly higher accuracy in capturing the input-output dynamics than TeaCache.}
\label{fig.all fit}
\end{figure*}

\begin{table}[htbp]
\centering
\caption{Table shows the mean squared error (MSE) values, with units of $10^{-4}$, obtained by comparing the output curves of Teacache and PromptTea algorithms against a set of labels. It is evident that PromptTea has smaller MSE values, indicating a more precise fit.}
\begin{tabular}{c|ccc}
\toprule[2pt]
Method & CogVideoX1.5 & HunyuanVideo & Wan2.1 \\
\midrule
TeaCache & 84.51 & 1.67 & 82.62 \\
PromptTea & 2.85 & 0.38 & 2.55 \\
\bottomrule[2pt]
\end{tabular}
\label{table.mse}
\end{table}

\subsubsection{Input-Output Correlation Analysis}
As illustrated in \text{Fig. \ref{fig.all diff2}}, this section examines the input-output correlations across three models: CogVideoX1.5, HunyuanVideo, and Wan2.1. Empirical results reveal that timestep embedding demonstrates a stronger correlation with output than TEMNI across all models—a finding that leads TeaCache to adopt timestep embedding difference as the input feature for estimating output difference in Wan2.1 and CogVideoX1.5.

Notably, several anomalies challenge the universality of assumptions regarding input-output correlations. For CogVideoX1.5, timestep embedding difference increases over the first three timesteps, yet the corresponding output difference decreases. In Wan2.1, timestep embedding difference remains nearly constant during the first five timesteps, while output difference undergoes a sharp decline. These discrepancies indicate that the earlier assumption of a strong input-output connection does not hold universally.

\subsubsection{Specific Examples of Discrepancies}
Further evidence of these inconsistencies emerges from comparative analyses of specific timesteps. In CogVideoX1.5, for instance, the timestep embedding differences between the 1st and 12th timesteps are comparable in magnitude, yet their corresponding output differences diverge significantly. Similarly, in Wan2.1, the timestep embedding differences at the 1st and 19th timesteps are both approximately 0.009, but the associated output differences are starkly distinct (0.63 vs. 0.053). These cases demonstrate that relying exclusively on timestep embedding differences to predict output differences would introduce substantial errors, as identical input values (e.g., 0.009) can map to vastly disparate outputs (e.g., 0.63 or 0.053).

\subsubsection{Incorporating Positional Information}
Drawing inspiration from diffusion models, we propose a novel solution: integrating positional information into input features (see \text{Eq. \ref{eq.T and t}}). This enables the model to pinpoint its current stage in the sequence, facilitating more accurate fitting of input-output relationships. Notably, this approach effectively addresses a key limitation of TeaCache—specifically, its predictions often exhibited monotonic trends inconsistent with actual output behavior (e.g., rising predictions versus declining ground truth in early timesteps).

\subsubsection{Validation through MSE Comparison}
\text{Tab. \ref{table.mse}} reports the Mean Squared Error (MSE) between predicted output curves (from our improved PromptTea algorithm and the baseline TeaCache) and ground truth labels. PromptTea exhibits significantly lower MSE across all models, directly validating the effectiveness of our proposed improvements.

\subsection{Uniform CFGCache}
Based on the observation that there is significant redundancy between conditional and unconditional outputs at the same timestep in diffusion models, FasterCache 
\cite{fastercache} proposes a frequency - aware uniform cache strategy - - CFGCache to reduce computational overhead. This method decomposes the difference between these outputs into frequency components through Fast Fourier Transforms (FFT):
\begin{equation}
    \Delta_{LF} = \mathcal{FFT}(\epsilon_{\theta}(x, t, \emptyset))_{low} - \mathcal{FFT}(\epsilon_{\theta}(x, t, c))_{low}
    \label{eq.lf}
\end{equation}
\begin{equation}
    \Delta_{HF} = \mathcal{FFT}(\epsilon_{\theta}(x, t, \emptyset))_{high} - \mathcal{FFT}(\epsilon_{\theta}(x, t, c))_{high}
    \label{hf}
\end{equation}

In these equations, $\mathcal{FFT}(\epsilon_{\theta}(x, t, c))$ represents the conditional output (predicted noise) at timestep $t$ with guidance $c$, and $\mathcal{FFT}(\epsilon_{\theta}(x, t, \emptyset))$ represents the unconditional output. The subscripts $low$and $high$ denote low - frequency and high - frequency components obtained through Fast Fourier Transform (FFT). $\Delta L_F$ and $\Delta H_F$ capture the frequency - domain biases between conditional and unconditional outputs.

These biases are cached and adaptively applied in subsequent timesteps. For example, in the inference process at timestep $t - i$, the unconditional output is reconstructed as follows:
\begin{align}
    \epsilon_{\theta}(x_{t-i}, t-i, \emptyset) =& \mathcal{IFFT}\bigl(  \mathcal{FFT}(\epsilon_{\theta}(x_{t-i}, t-i, c)) \nonumber \\
    &+ w_1 \cdot \Delta L_{F} + w_2 \cdot \Delta H_{F} \bigr)
    \label{eq.uniform cfgcache}
\end{align}    

Here, $\mathcal{IFFT}$ is the Inverse Fast Fourier Transform, and $w_1$ and $w_2$ are adaptive weights controlled by timestep $t$. In the early stages ( $t < t_0$), low - frequency components are prioritized, and in the later stages ($t \geq t_0$), high - frequency components are prioritized. This method balances computational efficiency and generation quality by selectively reusing cached frequency - domain components.

\subsection{Differences Between PromptTea and Other Algorithms}
\begin{figure*}[htbp]
    \centering
    \includegraphics[width=0.8\linewidth]{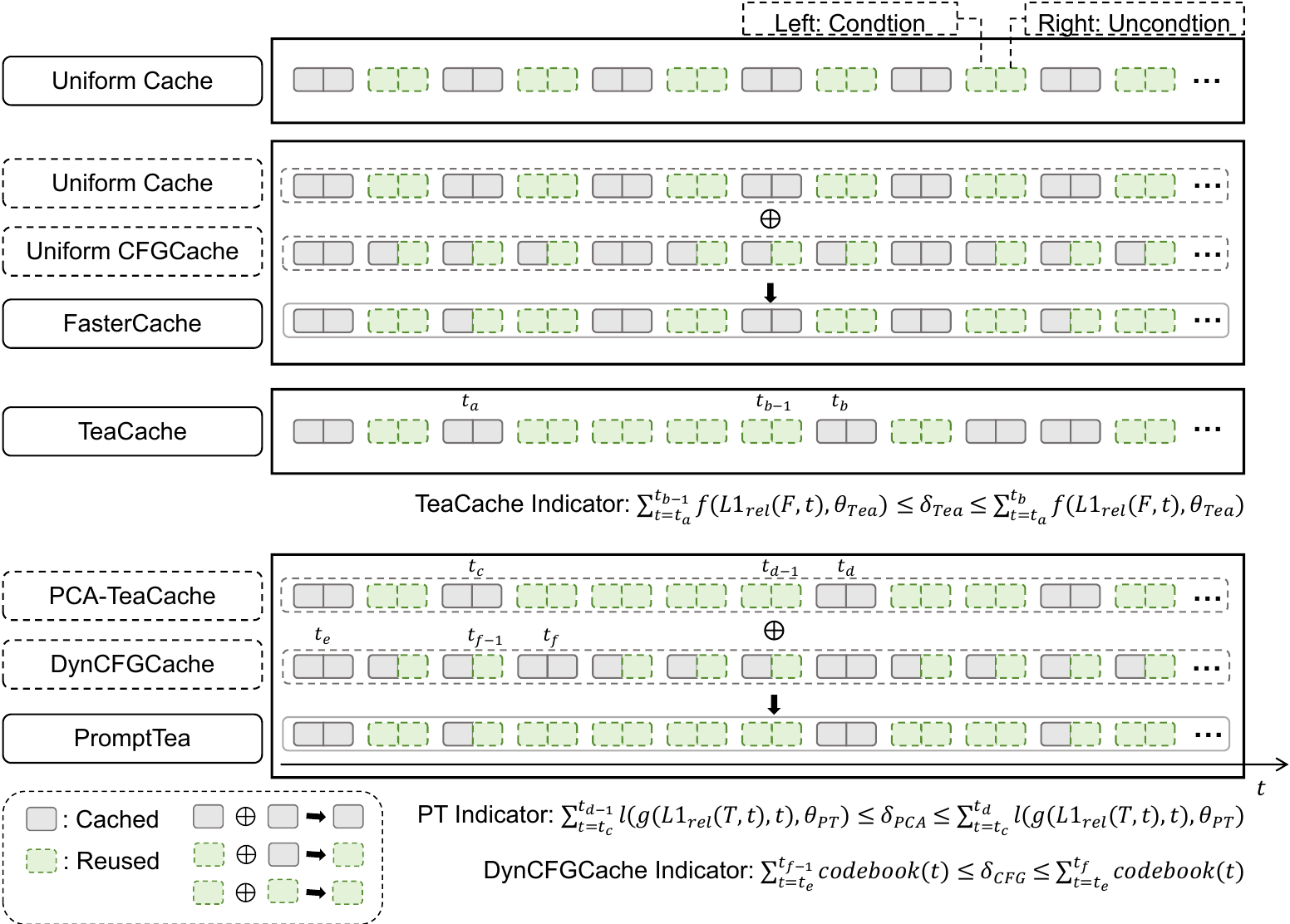}
    \caption{Uniform Cache reuses cache at fixed-interval steps. FasterCache extends Uniform Cache by adding cache reuse for CFG, also using fixed intervals in the CFG dimension. TeaCache determines cache reuse by comparing the estimated cumulative output difference against a manually set threshold. PromptTea’s PCA-TeaCache evolves from TeaCache via input-output relationship reconstruction and PCA-driven automated threshold setting. Beyond PCA-TeaCache, PromptTea introduces DynCFGCache, an enhanced CFGCache that replaces fixed-interval reuse with a TeaCache-like mechanism—comparing estimated cumulative output differences to thresholds—for CFG-based cache reuse. Other details are consistent with \text{Fig. \ref{fig.toc}}.}
    \label{fig.toc2}
\end{figure*}

As illustrated in \text{Fig. \ref{fig.toc2}}, within the domain of caching mechanisms, uniform cache stands as a prevalent type that reuses cached data at fixed-interval steps. FasterCache extends this uniform caching paradigm by introducing fixed-interval cache reuse for CFG (Classifier-Free Guidance) in the CFG dimension. In contrast, TeaCache determines cache reuse through a threshold-based mechanism, where the estimated cumulative output difference is compared against a manually set threshold.

PromptTea’s PCA-TeaCache evolves from TeaCache via two key enhancements: reconstructed input-output relationships and PCA-driven automated threshold setting. Beyond PCA-TeaCache, PromptTea introduces DynCFGCache—an enhanced variant of CFGCache. Departing from fixed-interval reuse, DynCFGCache adopts a TeaCache-like strategy for CFG-based cache reuse, where decisions are made by comparing estimated cumulative output differences against thresholds.

Notably, Uniform CFGCache and DynCFGCache exhibit fundamental distinctions. Uniform CFGCache employs a frequency-domain perception strategy: it decomposes differences between conditional and unconditional outputs into low-frequency and high-frequency components using Fast Fourier Transform (FFT), then selectively reuses these frequency-domain biases across stages with fixed adaptive weights. DynCFGCache, by contrast, abandons frequency-domain decomposition entirely, instead relying on dynamic cumulative difference assessment.

As shown in \text{Tab. \ref{table.fft}}, incorporating FFT yields negligible differences in objective metrics. The faster performance of FFT-based methods stems from their direct reuse of cached video latent representations, whereas our approach operates on sequential latents—introducing additional steps for video-to-sequence conversion and sequence-to-video reconstruction, which inherently increase latency. While FFT integration improves VBench2 scores, other visual quality metrics underperform compared to the non-FFT baseline. After comprehensive evaluation, we opt not to implement FFT in our framework.

\begin{table*}[htbp]
\centering
\caption{Ablation experiments of FFT modules.}
\begin{tabular}{c|ccccc}
\toprule[2pt]
& Latency (s) $\downarrow$ & VBench2 $\uparrow$ & LPIPS $\downarrow$ & SSIM $\uparrow$ & PSNR $\uparrow$ \\
\midrule
PromptTea(0.1, 0.23) & 674.46 & 0.4710 & 0.1380 & 0.7884 & 23.00 \\
PromptTea(0.1, 0.23)+FFT & 660.73 &  0.4768 & 0.1425 & 0.7837 & 22.90 \\
\bottomrule[2pt]
\end{tabular}
\label{table.fft}
\end{table*}

\subsection{Conflict between Text-Guidance Distillation and CFGCache}
HunyuanVideo's text-guidance distillation fuses unconditional and conditional inputs into a single student model during training, merging their separate computations. This eliminates the distinction between conditional/unconditional paths, making CFGCache's reuse of conditional cache for unconditional inference impossible, hence the conflict.

\subsection{Visualization}
\begin{figure*}[htbp]
    \centering
    \includegraphics[width=1\linewidth]{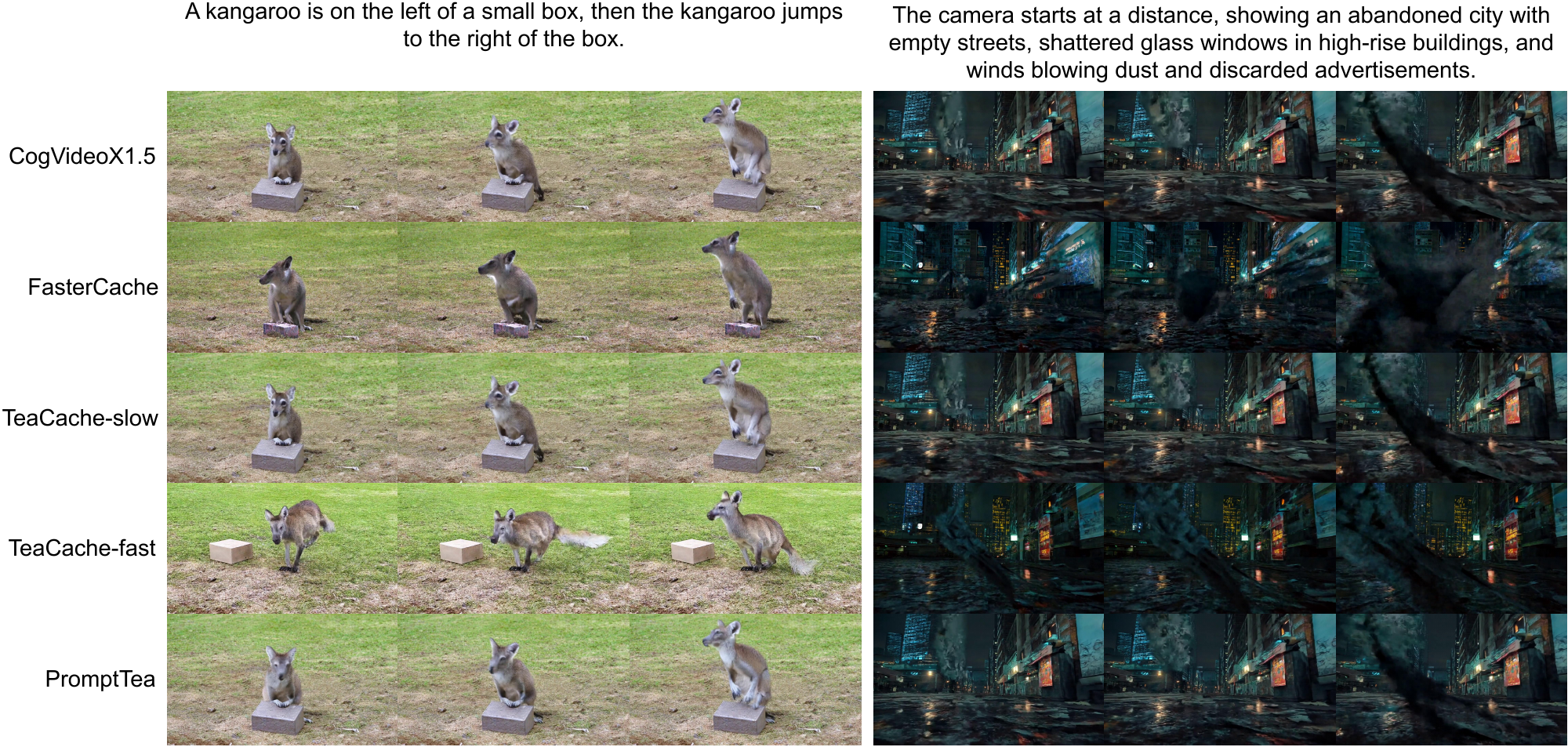}
    \caption{Visualization results for different acceleration methods on CogVideoX1.5 model}
    \label{fig.CogVideoX1.5 Visual}
\end{figure*}

\begin{figure*}[htbp]
    \centering
    \includegraphics[width=1\linewidth]{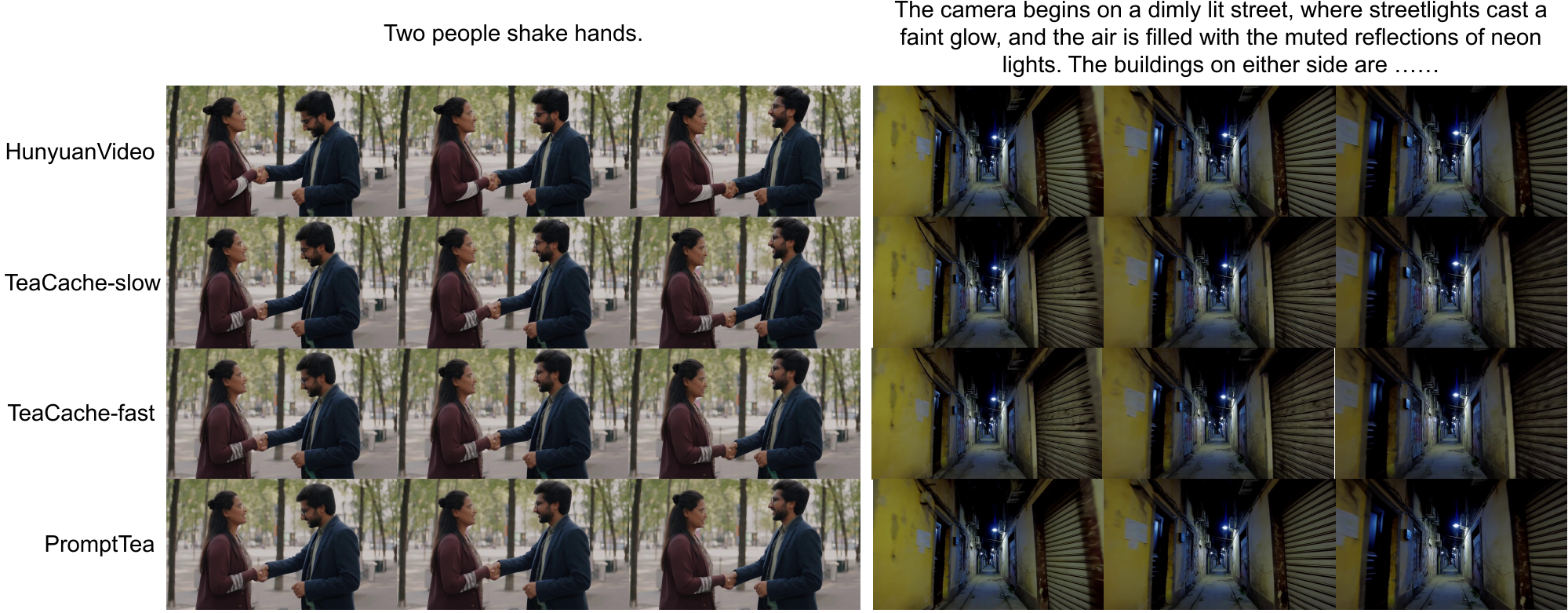}
    \caption{Visualization results for different acceleration methods on HunyuanVideo model}
    \label{fig.HunyuanVideo Visual}
\end{figure*}

\begin{figure*}[htbp]
    \centering
    \includegraphics[width=1\linewidth]{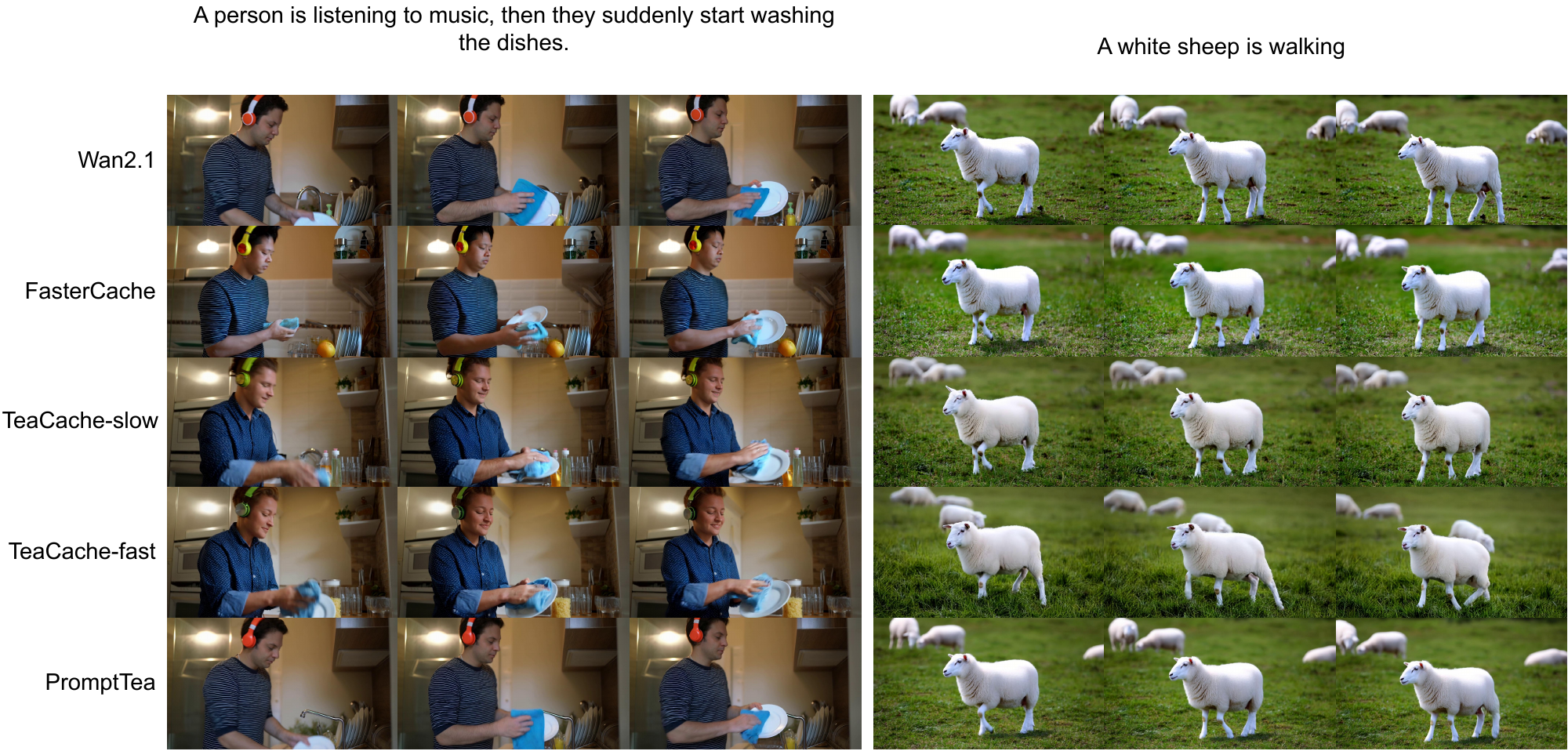}
    \caption{Visualization results for different acceleration methods on Wan2.1 model}
    \label{fig.Wan2.1 Visual}
\end{figure*}

\begin{figure*}[htbp]
    \centering
    \includegraphics[width=1\linewidth]{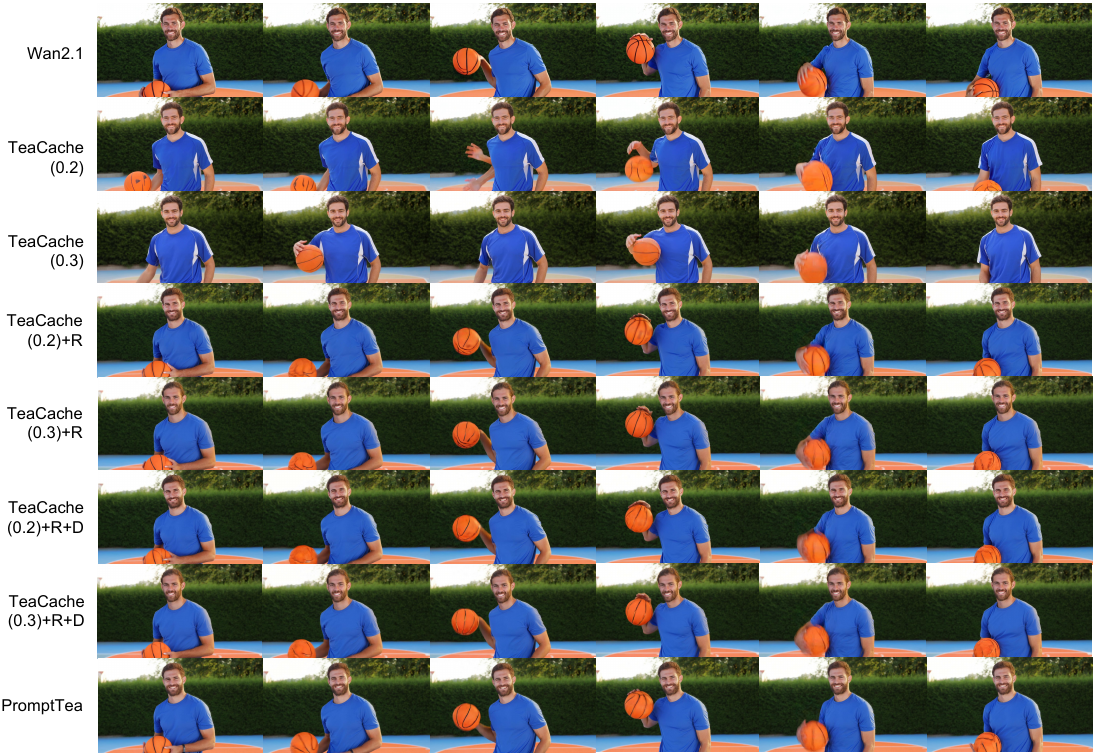}
    \caption{Ablation study visual comparison}
    \label{fig.ablation vsual}
\end{figure*}

The visual comparison figures of the experiments, including the comparative experiments and ablation experiments, are shown in Figs. \ref{fig.CogVideoX1.5 Visual}-\ref{fig.ablation vsual}.

\end{document}